\documentclass[nohyperref]{article}

\usepackage{microtype}
\usepackage{graphicx}
\usepackage{subfigure}
\usepackage{booktabs} % for professional tables

\usepackage{hyperref}

\usepackage[accepted]{icml2022}

\usepackage{amsmath}
\usepackage{amssymb}
\usepackage{mathtools}
\usepackage{amsthm}
\usepackage{listings}
\usepackage{xcolor}

\definecolor{codegreen}{rgb}{0,0.6,0}
\definecolor{codegray}{rgb}{0.5,0.5,0.5}
\definecolor{codepurple}{rgb}{0.58,0,0.82}
\definecolor{backcolour}{rgb}{0.95,0.95,0.92}

\lstdefinestyle{mystyle}{
    backgroundcolor=\color{backcolour},   
    basicstyle=\ttfamily\footnotesize,
    breaklines=true,
    captionpos=b,
    keepspaces=false,
    breakindent=0pt
}

\usepackage[capitalize,noabbrev]{cleveref}

\theoremstyle{plain}

\theoremstyle{definition}

\theoremstyle{remark}

\usepackage[textsize=tiny]{todonotes}

\icmltitlerunning{}

\begin{document}

\twocolumn[
\icmltitle{Stochastic Code Generation}



\icmlsetsymbol{equal}{*}

\begin{icmlauthorlist}
\icmlauthor{Swapnil Sharma}{yyy}
\icmlauthor{Nikita Anand}{yyy}
% \icmlauthor{Firstname3 Lastname3}{comp}
\icmlauthor{Kranthi Kiran GV}{yyy}
\end{icmlauthorlist}

\icmlaffiliation{yyy}{Department of Computer Science, Courant Institute of Mathematical Sciences, New York University}
%\icmlaffiliation{comp}{Company Name, Location, Country}
%\icmlaffiliation{sch}{School of ZZZ, Institute of WWW, Location, Country}

\icmlcorrespondingauthor{Swapnil Sharma}{ss14412@nyu.edu}
\icmlcorrespondingauthor{Nikita Anand}{lna285@nyu.edu}
\icmlcorrespondingauthor{Kranthi Kiran GV}{kranthi.gv@nyu.edu}

\icmlkeywords{Machine Learning, ICML}

\vskip 0.3in
]



\printAffiliationsAndNotice{}  % leave blank if no need to mention equal contribution

\begin{abstract}
Large language models pre-trained for code generation can generate high-quality short code, but often struggle with generating coherent long code and understanding higher-level or system-level specifications. This issue is also observed in language modeling for long text generation, and one proposed solution is the use of a latent stochastic process. This approach involves generating a document plan and then producing text that is consistent with it.

In this study, we investigate whether this technique can be applied to code generation to improve coherence. We base our proposed encoder and decoder on the pre-trained GPT-2 based CodeParrot model and utilize the APPS dataset for training. We evaluate our results using the HumanEval benchmark and observe that the modified Time Control model performs similarly to CodeParrot on this evaluation.
% add short summary of results
\end{abstract}

\section{Introduction}
%generic intro to nlp tasks 
Large pre-trained language models have gained immense popularity recently. Large language models have achieved impressive results in various natural language processing tasks, including language translation, summarization, and text generation. These models are trained on a large corpus of text data and are able to capture complex patterns and relationships in the data. One impactful application of these models is code generation.

%intro to code gen
Code generation is the process of automatically generating computer programs based on natural language descriptions or higher-level specifications. Multiple large scale models like PyMT5 \cite{clement-etal-2020-pymt5}, Codebert \cite{DBLP:journals/corr/abs-2002-08155} and Codex \cite{chen2021codex} have been trained for the task and are used to aid software developers. They are used in integrated development environments for code completion, to convert natural language documentation (docstrings) into code and vice versa.

%incoherency in code gen
These models produce high quality short pieces of code but as pointed out by \citealp{chen2021codex}, they tend to struggle with long and higher-level or system-level specifications. An experiment was run on Codex that indicates performance degradation as docstring length increases. 

%briefly- incoherency in text gen and time control paper
As stated by \citealp{wang2022language}, this problem also occurs in text generation where models tend to meander or lose coherency when producing long text. They propose Time Control, a language model that learns a mapping of how text changes in a document with the changes in the document plan generated via a stochastic process. Text is then generated to be consistent with this document plan. This helps the model learn structure and leads to locally and globally more coherent text.

%adapting to code gen
In this work, we adapt the stochastic process for the code generation task. We do this by basing the proposed encoder-decoder architecture on the pre-trained CodeParrot model \citep{tunstall2022natural} which was trained using 180 gigabytes of Python code. We use the Automated Programming Progress Standard dataset \citealp{hendrycksapps2021} to generate the latent plan and train the encoder and fine-tune the decoder.

%evaluation
We compare our model's (Modified Time Control) performance to the regular pre-trained CodeParrot model on the HumanEval \citep{chen2021codex} Evaluation framework and see that they perform similarly. We also compare the results manually to get a better understanding of when it works well and when it doesn't.

% add short summary of results

\section{Related Work}
Autoregressive models, which are commonly used for generating text, may struggle to produce long and coherent text because they lack the ability to model text structure and dynamics \cite{wang2022language, lin2021limitations}. This can result in globally incoherent text as the models are unable to effectively plan and anticipate long context in the text. When forced to generate longer texts, this incoherence is often worsened as the models struggle to extend beyond their expected text end point. To address this issue, previous research has employed planning-based approaches in an attempt to generate globally coherent text \cite{puduppully2019data, kiddon2016globally}.

Program induction is a machine learning technique in which a model generates program outputs from a latent program representation \cite{zaremba2014learning, pierrot2019learning}. Program synthesis involves generating a program from a natural language specification \cite{feng2020codebert}. Both approaches have been improved upon through the incorporation of inductive biases and the use of recurrence and abstract syntax trees.

The Codex paper by \citealp{chen2021codex} introduces a large language model called Codex that is finetuned on code from GitHub and is able to generate Python code. The model was evaluated on the HumanEval dataset \citep{chen2021codex}, which measures functional correctness for synthesizing programs from docstrings, and was found to solve 28.8\% of the problems, outperforming the GPT-3 model which solved 0\%. The limitations of Codex include difficulty with docstrings describing long chains of operations and with binding operations to variables.

\section{Method}
In this section we describe the parts of \textit{Language Modeling via Stochastic Processes} \citep{wang2022language} that we use to run our experiments. In the following Experiments section, we describe the changes made to adapt this method to code generation in detail. 

\subsection{Architecture}
We adapt the same structure as Time Control but make modifications to run for code generation. The proposed solution is a modified encoder-decoder model built on top of CodeParrot.

\paragraph{Encoder}
The encoder in this study is a nonlinear mapping that transforms raw input data into a low-dimensional latent space. The goal of the encoder is to map high-dimensional sequential data into a latent space that follows a stochastic process of interest, similar to the Brownian bridge process described by \citealp{wang2022language}. 

As demonstrated by \citealp{wang2022language}, the density of a Brownian bridge process between an arbitrary start point, $z_0$, at time $t=0$ and end point, $z_T$, at time $t=T$, is given by:

$$ p(z_t|z_0,z_T) = \mathcal{N}\left(\left(1-\frac{t}{T}\right)z_0+\frac{t}{T}z_T, \frac{t(T-t)}{T}\right) $$

This equation acts as a noisy interpolation between the start and end points of the trajectory, with the center being the most noisy. To ensure that every triplet of observations, ($x_1, x_2, x_3$), follows the Brownian bridge transition density, we use a contrastive objective.

\paragraph{Decoder}
The decoding component of the system is responsible for generating code from latent plans. To do this, we first map all lines in the training dataset to our learned latent space using the pretrained encoder, $f_\theta$. This produces a Brownian bridge trajectory of sentence-level latent codes, $(z_0,\cdots,z_t,\cdots,z_T)$, for each entry in the dataset. Rather than training a decoder from scratch, we fine-tune the existing model, CodeParrot, to generate code conditioned on past context and the latent structure.

To fine-tune CodeParrot, we use a standard auto-regressive language model that has been modified as follows. Given a document with $W$ tokens and $T$ sentences used to train the decoder, $x_1\cdots x_W$, we obtain embeddings, $z_1\cdots z_3$, for each sentence using the encoder, $f_\theta$. At time $t$, the decoder must predict $x_t$ using all past tokens, $x_{<t}$, as well as the sentence embedding, $z_{s_t}$, where the index $s_t \in [T]$ maps each token to its corresponding sentence. This process can be viewed as a reconstruction objective, as the identity of $x_t$ is encoded in $z_{s_t}$.

\begin{figure*}
\begin{lstlisting}[caption=Sample Introductory Level Question from APPS]
-----Question-----

Polycarp analyzes the prices of the new berPhone. At his disposal are the prices for n last days: a_1, a_2, ... , a_n, where a_i is the price of berPhone on the day i. Polycarp considers the price on the day i to be bad if later (that is, a day with a greater number) berPhone was sold at a lower price. For example, if n=6 and a=[3, 9, 4, 6, 7, 5], then the number of days with a bad price is 3 - these are days 2 (a_2=9), 4 (a_4=6) and 5 (a_5=7). Print the number of days with a bad price. You have to answer t independent data sets. 


-----Input-----

The first line contains an integer t (1 <= t <= 10000$) - the number of sets of input data in the test. Input data sets must be processed independently, one after another. Each input data set consists of two lines. The first line contains an integer n (1 <= n <= 150000) - the number of days. The second line contains n integers a_1, a_2, ... , a_n (1 <= a_i <= 10^6), where a_i is the price on the i-th day. It is guaranteed that the sum of n over all data sets in the test does not exceed 150000.

-----Output-----

Print t integers, the j-th of which should be equal to the number of days with a bad price in the j-th input data set.

-----Example-----

Input 5 6 3 9 4 6 7 5 1 1000000 2 2 1 10 31 41 59 26 53 58 97 93 23 84 7 3 2 1 2 3 4 5
Output 3 0 1 8 2
\end{lstlisting}
\end{figure*}

\begin{table}[t]
\label{sample-table}
\vskip 0.15in
\begin{center}
\begin{small}
\begin{sc}
\begin{tabular}{lcccr}
\toprule
Level & Total Count & Test Set \\
\midrule
Introductory  & 3639 &  1000 \\
Interview & 5000 & 3000\\
Competition    & 1361  & 1000\\
\bottomrule
\end{tabular}
\end{sc}
\end{small}
\end{center}
\caption{Problems of varying difficulty levels in the APPS dataset}
\vskip -0.1in
\end{table}

\subsection{Dataset}
The Automated Programming Progress Standard (APPS) dataset introduced by \citealp{hendrycksapps2021} consists of coding problems collected from multiple open access websites like Codeforces, Kattis etc. It contains 10,000 problems of varying levels; introductory, interview level and competition level. The introductory problems are ones whose solutions do not require any complicated algorithms, and can be solved by people with a year or two of programming experience. The interview level questions are ones that are used to test technical proficiency and require non-trivial solutions. They can typically involve the use of data structures like trees, linked lists etc. The competition level questions are ones that require advanced collegiate level knowledge.

The latent structure was labelled using an open-source tool called tree-sitter\footnote{\href{https://github.com/tree-sitter/tree-sitter}{https://github.com/tree-sitter/tree-sitter}}. This tool is a parser that builds a syntax tree for a source file. It works with any programming language, is robust enough that it can work with minor errors and can be embedded in any application because it is dependency free. 

\subsection{Choice of Model}
Some of the most popular language models like Codex which powers the industry product Github Copilot, are not open-source and thus unusable in this work. Our choice of pre-trained model for code generation is largely based on what was open sourced and popular on the Hugging Face library \citep{DBLP:journals/corr/abs-1910-03771}. 

For our experiments we use CodeParrot \citep{tunstall2022natural} which is a GPT-2 \citep{radford2019language} based model trained to generate Python code. It was downloaded 22,416 times from Hugging Face in the last month. CodeParrot is an auto-regressive Left-to-Right Language Model that predict the probability of a token given the previous tokens. It has 1.5 billion parameters and was trained on 180 GB of Python code. The Time Control model \citep{wang2022language} we reference is also based on GPT-2 which lead us to believe this model would be a good fit for our experiments.

\section{Experiments}
The purpose of this study is to evaluate the capabilities of Time Control in capturing code structure and generating more robust code. Specifically, we aim to determine whether the generated code is able to compile and solve the problem stated in the prompt. To assess Time Control, we use a modified version with latent dimensions (d = 32).

The encoder architecture for this evaluation consists of a frozen, pretrained CodeParrot and a trainable multi-layer perceptron (MLP). We extract the last layer hidden state of CodeParrot that corresponds to the end-of-sentence token, and train the four-layer MLP on top of this hidden state. The MLP has intermediate rectified linear unit activations and is trained using stochastic gradient descent with a learning rate of 1e-4 and momentum of 0.9.

For the decoder, we also use CodeParrot as the base model. Similar to previous work \cite{wang2022language}, we modify CodeParrot to pass the encoder output after appending it to positional embeddings in the transformer layer. You can see one of the good samples generated in Figure \ref{fig:sample-gen-code} and some of the bad ones \ref{fig:bad-sample-gen-code}.

\begin{table}[t]
\caption{Model accuracy with \textit{baseline} CodeParrot.}
\label{human-eval}
\vskip 0.15in
\begin{center}
\begin{small}
\begin{sc}
\begin{tabular}{lccr}
\toprule
Model & Pass@1 & Pass@10 & Pass@100 \\
\midrule
TimeControl & 3.7\% & 6.58\% & 12.73\% \\
\midrule
CodeParrot 110M & 3.80\% & 6.57\% & 12.78\% \\
CodeParrot 1.5B & 3.58\% & 8.03\%	& 14.96\% \\
\bottomrule
\end{tabular}
\end{sc}
\end{small}
\end{center}
\vskip -0.1in
\end{table}
\paragraph{Datasets} As the dataset for this study, we use APPS, which provides sufficient length code to fully utilize our available resources. We modify the dataset so that each entry contains five sections (question, solution, class statement, def statement, import statement) marked with section identifier tokens. Specifically, each entry is represented as,
\begin{verbatim}
    [QUESTION]
        <text>
        
    [SOLUTION] 
        [CLASS_STATEMENT]
            <code>
        [DEF_STATEMENT]
            <code>
        [IMPORT_STATEMENT]
            <code>
\end{verbatim}
No additional tokens are added in this dataset.

To label the code in the dataset with the appropriate section identifiers, we utilize tree-sitter to mark each line of code with tree-sitter specific labels, and then retain only the relevant labels while discarding the rest. 

\begin{figure*}
    \begin{lstlisting}
    [QUESTION]:
    In this problem we assume the Earth to be a completely round ball and its surface a perfect sphere. The length of the equator and any meridian is considered to be exactly 40 000 kilometers. Thus, travelling from North Pole to South Pole or vice versa takes exactly 20 000 kilometers.
    
    Limak, a polar bear, lives on the North Pole. Close to the New Year, he helps somebody with delivering packages all around the world. Instead of coordinates of places to visit, Limak got a description how he should move, assuming that he starts from the North Pole. The description consists of n parts. In the i-th part of his journey, Limak should move t_{i} kilometers in the direction represented by a string dir_{i} that is one of: "North", "South", "West", "East".
    
    Limak isn't sure whether the description is valid. You must help him to check the following conditions:  If at any moment of time (before any of the instructions or while performing one of them) Limak is on the North Pole, he can move only to the South . If at any moment of time (before any of the instructions or while performing one of them) Limak is on the South Pole, he can move only to the North . 
    
    [SOLUTION]:
    def main():
    	latitude = 0
    	n, = read()
    	for i in range(n):
    		l, d = read(str)
    		l = int(l)
    		if latitude == 0:
    			if d != "South":
    				return "NO"
    		if latitude == 20000:
    			if d != "North":
    				return "NO"
    		if d == "South":
    			latitude += l
    		if not (0 <= latitude <= 20000):
    			return "NO"
    	if latitude != 0:
    		return "NO"
    	return "YES"
    
    def read(callback=int):
    	return list(map(callback, input().strip().split()))
    
    def write(value, end="\n"):
    	if value is None: return
    	try:
    		if not isinstance(value, str):
    			value = " ".join(map(str, value))
    	except:
    		pass
    	print(value, end=end)
    write(main())
    \end{lstlisting}
    \caption{Problem Statement and Solution generated by Modified Time Control}
    \label{fig:sample-gen-code}
\end{figure*}

\begin{figure*}
    \begin{lstlisting}
    [QUESTION]
    Given an array of is, you can find a array where each element has equal letters.
    
    [SOLUTION]
    a = list(str(input()))
    for i in range(1,len(a)):
     sh = a[i].swapcase()
     tell = sh['what']
      ...
    x,n = list(a.swapcase())  a.sort()  pivot = """"  pivot_pos = pivot_pos__starting_point__starting_point__starting_point__starting_point__len()  if pivot_pos:   pivot_pos += 1      pivot_pos = pivot_pos__starting_point__starting_point__end() gcdrom = pivot ocdr = pivot while True:  a,r = pivot_pos__starting_point__starting_point__starting_point...
    \end{lstlisting}
    \caption{Sample from Modified Time Control with incomprehensible question and output}
    \label{fig:bad-sample-gen-code}
\end{figure*}

\subsection{Evaluation}
To evaluate the model, we use the HumanEval framework. HumanEval contains 164 handwritten programming problems; this is done to ensure that the models could not have seen the problems before as most are trained on open access code from Github which contain solutions to problems from the web like the APPS dataset. 

We don't use metrics like BLEU Score that are used for natural language because it neglects syntactic and semantic features of code \citep{DBLP:journals/corr/abs-2009-10297} and doesn't work well for evaluatinh code generation.

Our results, shown in Table \ref{human-eval}, indicate that modified Time Control performs similarly to CodeParrot on the HumanEval dataset. As described by \citealp{chen2021codex}, "Pass@k can be interpreted as the result of evaluating the best out of k samples, where the best sample is picked
by an oracle with prior knowledge of the unit tests". These findings suggest that Modified Time Control is a promising approach, as we were able to maintain code quality with our limited computational resources. We hope that with longer per-sentence lengths and additional computing resources, we will be able to further improve the performance of our model.

\section{Conclusion}
The purpose of this study was to investigate the potential of Time Control to improve code generation through the use of a latent stochastic process that generates output with learned implicit structure. Our results suggest that modified Time Control is capable of learning to map programming code to smooth Brownian bridge trajectories, and in some cases, we observed improved performance in the code generation process. These findings are encouraging for the use of Time Control in code generation.

To further explore the potential of Time Control, we suggest using larger datasets containing multiple files in a project as dataset entries. We believe that this approach may allow for the learning of code-level architecture. In addition, the use of enhanced datasets and additional computing resources may lead to even more improved results with modified Time Control.

\section{Future Work}
Due to computational constraints, we were unable to utilize approximately 70\% of our training set and had to limit the decoder to less than 750 tokens. It is possible that the use of the full dataset may have produced different results. 

In an ideal scenario, we would have preferred to conduct these experiments using a more structured dataset, such as programs with the model-view-controller software architecture pattern. Such a dataset would provide higher-level specifications and have a well-defined structure that may be particularly well-suited to benefit from a latent process like the one used in this study.

\bibliography{example_paper}
\bibliographystyle{icml2022}

%%%%%%%%%%%%%%%%%%%%%%%%%%%%%%%%%%%%%%%%%%%%%%%%%%%%%%%%%%%%%%%%%%%%%%%%%%%%%%%
%%%%%%%%%%%%%%%%%%%%%%%%%%%%%%%%%%%%%%%%%%%%%%%%%%%%%%%%%%%%%%%%%%%%%%%%%%%%%%%
% APPENDIX
%%%%%%%%%%%%%%%%%%%%%%%%%%%%%%%%%%%%%%%%%%%%%%%%%%%%%%%%%%%%%%%%%%%%%%%%%%%%%%%
%%%%%%%%%%%%%%%%%%%%%%%%%%%%%%%%%%%%%%%%%%%%%%%%%%%%%%%%%%%%%%%%%%%%%%%%%%%%%%%
% \newpage
% \appendix
% \onecolumn

% You can have as much text here as you want. The main body must be at most $8$ pages long.
% For the final version, one more page can be added.
% If you want, you can use an appendix like this one, even using the one-column format.
%%%%%%%%%%%%%%%%%%%%%%%%%%%%%%%%%%%%%%%%%%%%%%%%%%%%%%%%%%%%%%%%%%%%%%%%%%%%%%%
%%%%%%%%%%%%%%%%%%%%%%%%%%%%%%%%%%%%%%%%%%%%%%%%%%%%%%%%%%%%%%%%%%%%%%%%%%%%%%%

\end{document}